# Can DMD obtain a Scene Background in Color?

Santosh Tirunagari, Norman Poh, Miroslaw Bober and David Windridge*
Faculty of Engineering and Physical Sciences, University of Surrey, Guildford, UK.
*Department of Computer Science, Middlesex University, Hendon, London.
E-mail: {s.tirunagari, n.poh, m.bober}@surrey.ac.uk & d.windridge@mdx.ac.uk

*Abstract—* A background model describes a scene without any foreground objects and has a number of applications, ranging from video surveillance to computational photography. Recent studies have introduced the method of Dynamic Mode Decomposition (DMD) for robustly separating video frames into a background model and foreground components. While the method introduced operates by converting color images to grayscale, we in this study propose a technique to obtain the background model in the color domain. The effectiveness of our technique is demonstrated using a publicly available Scene Background Initialisation (SBI) dataset. Our results both qualitatively and quantitatively show that DMD can successfully obtain a colored background model.

## I. INTRODUCTION

The objective of Scene Background Initialisation (SBI) is to obtain a background model from a sequence of images where the background is occluded with a number of foreground objects. It has a number of applications, including video surveillance, video segmentation, video compression, video inpainting, privacy protection for videos, and computational photography [1].

A recent study by Maddalena *et al.* [2], [3] reviewed and benchmarked five different methods that are suitable for obtaining a background model from a given image sequence, namely, temporal median (baseline) method, Spatially Coherent Self-Organizing Background Subtraction (SC-SOBS) [4], WS2006 [5], RSL2011 [6] and Photomontage [7]. In the temporal median method the background model is obtained as the median of pixel values at the same location throughout the image sequence. SC-SOBS estimates the background model by detecting moving foreground objects using a self-organizing neural background model. WS2006 initialises the background model in a two-step process. First, for each pixel, the longest stable sequence of values that have similar intensities in the image sequence is considered as a candidate background. Second, using a RANSAC method the stable sequences which are likely to arise from the background are selected. The temporal mean of the selected subsequence provides the estimated background model. For RSL2011 the background estimation is carried out at the image block level using Markov random fields. A combined frequency is calculated on an image block and its corresponding neighbourhood. The background model is then estimated from the blocks that provide the smoothest frequency from the image sequence. Finally, the Photomontage method initialises the background model through a framework that selects the image blocks that require minimum cost in editing images in the sequence.

In this paper, we will benchmark the recently introduced Dynamic Mode Decomposition (DMD) method, which robustly separates video frames into a background model and foreground objects [8], against the aforementioned list of methods. The DMD method decomposes a given image sequence into several images, called dynamic modes, which are associated with Fourier frequencies. The frequencies near the origin are interpreted as background modes of a given image sequence. In particular, DMD considers the parts in an image sequence that do not change in time as the background model. One of the key advantages of DMD is its data-driven nature which does not rely on any prior assumption about the form of objects in the image sequence except its inherent image dynamics which are captured in terms of 'DMD' modes. Although DMD was originally introduced in the area of computational fluid dynamics (CFD) [9], its capability for extracting relevant modes from complex fluid flows [10], [9], [11], [12], has gained significant applications in various fields [13], [14], [15], including for detecting spoof samples from facial authentication video datasets [16] and spoofed finger-vein images [17].

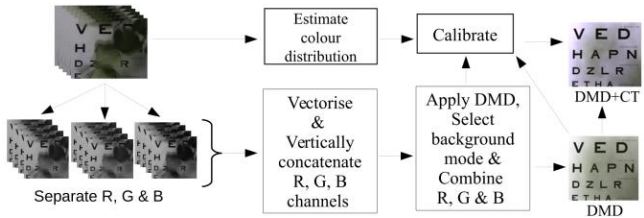

Fig. 1. Flow chart showing the steps involved in the methodological framework. A color image sequence consisting of $N$ images in R, G and B channels are vertically concatenated forming a data matrix $P^{RGB}$ as shown in the Equation 1. DMD is applied to this matrix to obtain $N-1$ dynamic mode images. Fourier frequencies are calculated on the dynamic eigenvalues. The mode with the Fourier frequency $\approx 0$ is then selected as the background initialised image. The three channels are then normalised and combined to obtain color background initialised image. Finally, the color is transferred from the original images to the obtained background image.

### A. Contribution

The DMD method that is introduced for obtaining the background model essentially requires the conversion of color images into grayscale images. However, it is sometimes necessary to have the scene background in the color domain, particularly for applications that include computational photography and video inpaintings. Therefore, our novel contribution in this paper is in modifying the DMD algorithm to obtain a color background model, thereby avoiding the conversion of color images to grayscale.

In this paper we answer the following questions:
1. Can DMD capture background models in the color domain?
2. How effective is the proposed DMD technique compared to existing benchmarked methods?

The remainder of the paper is as follows: In section II, we describe the SBI dataset, our methodological framework as well as implementation details. Results are discussed in section III. Finally, in section IV, we draw conclusions.

## II. DATA, METHODOLOGY & IMPLEMENTATION

In this section, we present the SBI dataset, our methodological framework as well as the implementation details.

### A. SBI dataset

The SBI [1] dataset includes seven image/bootstrap sequences along with their corresponding groundtruths (GT) as shown in Figure 2 (a & b). These sequences are challenging in their own way as described in [1].

### B. Methodology: DMD for color image sequences

Our methodological framework (Figure 1) consists of: (i) DMD applied to R, G and B channels of the color image sequence. (ii) Selection of the DMD mode representing the background model. (iii) Normalising and combining three channels of the selected DMD mode to obtain the colored background image. (iv) Transferring color from the original images to the obtained background image.

In a dynamic sequence of $N$ color images $P^{RGB}$, let $p_r$ be the $r^{th}$ image whose size is $m \times n \times 3$ i.e., with a height of m pixels and width of n pixels across the three color channels. This image $p_r$ is then separated for R, G and B channels. Later, these channels are vectorised to obtain a column vector of size $mn \times 1$ for each channel. These column vectors are then vertically concatenated to produce a vector of size $(3 \times mn) \times 1$ resulting in the construction of a data matrix $P^{RGB}$ of size $(3 \times mn) \times N$ for $N$ images in the sequence.

$$P^{RGB} = [p_1, p_2, p_3, \cdots, p_N] = \begin{pmatrix} R_1^1 & R_2^1 & \dots & R_N^1 \\ \vdots & \vdots & \vdots & \vdots \\ R_1^{mn} & R_2^{mn} & \dots & R_N^{mn} \\ G_1^1 & G_2^1 & \dots & G_N^1 \\ \vdots & \vdots & \vdots & \vdots \\ G_1^{mn} & G_2^{mn} & \dots & G_N^{mn} \\ B_1^1 & B_2^1 & \dots & B_N^1 \\ \vdots & \vdots & \vdots & \vdots \\ B_1^{mn} & B_2^{mn} & \dots & B_N^{mn} \end{pmatrix} \quad (1)$$

The images in the video sequence are collected over regularly spaced time intervals and hence each pair of consecutive images are linearly correlated. It can be justified that a linear mapping A exists between them forming a span of Krylov subspace [18], [19], [20]: P = $[p_1, Ap_1, A^2p_1, A^3p_1, \cdots, A^{N-1}p_1]$. The Krylov subspace can then be represented using two matrices $P_2$ and $P_1$ where $P_2 \equiv [p_2, p_3, \cdots, p_N]$ and $P_1 \equiv [p_1, p_2, \cdots, p_{N-1}]$.

$$P_2 = AP_1. \quad (2)$$

The mapping matrix $A$ is responsible for capturing the dynamics within the image sequence. The sizes of the matrices $P_2$ and $P_1$ are both $(3 \times mn) \times N-1$ each. Therefore, the size of the unknown matrix $A$ would be $(3 \times mn) \times (3 \times mn)$. Unfortunately, solving for A is computationally very expensive due to its size. For instance, if an image has a size of $240 \times 320 \times 3$ (Here 3 is denotes values from R, G and B channels) i.e., $m = 240$ and $n = 320$, the size of $A$ is then $230400 \times 230400$.

Since solving $A$ is computationally expensive for large image dimensions, we need an alternative solution. Our assumption that the images form a Krylov span, would allow us to introduce $H$.

$$AP_1 \approx P_1 H. \quad (3)$$

Here, $H$ is a companion matrix also known as a shifting matrix that simply shifts images 1 through $N - 1$ and approximates the last frame $N$ by linearly combining the previous $N - 1$ images, i.e., $p_N = c_0 p_1 + ... + c_N p_{N-1} = [p_1, p_2, p_3, \cdots, p_{N-1}]c$. $H$ requires the storage of $N-1 \times N-1$ data matrix which is significantly smaller than $A$ in dimensions.

$$H = \begin{pmatrix} 0 & 0 & \dots & 0 & c_0 \\ 1 & 0 & \dots & 0 & c_1 \\ & \ddots & \ddots & \vdots & \vdots \\ & & 1 & 0 & c_{N-2} \\ & & & 1 & c_{N-1} \end{pmatrix}. \quad (4)$$

Thus, for the last frame $N$, where $N$ is substantially lower dimensional than of $A$, one can write $P_2$ as a linear combination of the previous vectors. Consistent with Equations 2 and 3, we then have:

$$P_2 \approx P_1 H. \quad (5)$$

From Equations 3 and 5, we have $AP_1 \approx P_2 \approx P_1 H$.

In [9], the author describes a more robust solution, which is achieved by applying a singular value decomposition (SVD) on $P_1$. From Equation 2, the SVD decomposition on $P_1$ subspace is calculated to obtain $U$, $\Sigma$ and $V^*$ matrices that are left singular vectors, singular values and right singular vectors respectively. The inversions of these matrices are then multiplied with $P_2$ subspace to obtain the full-rank matrix $\tilde{H}$, determined on the subspace spanned by the orthogonal basis vectors $U$ of $P_1$, described by:

$$\tilde{H} = U^* P_2 V \Sigma^{-1}. \quad (6)$$

---

[1] http://sbmi2015.na.icar.cnr.it

Here, $U^* \in \mathbb{C}^{N \times M}$ and $V \in \mathbb{C}^{N \times N}$ are the conjugate transpose of $U$ and $V^*$, respectively; and $\Sigma^{-1} \in \mathbb{C}^{N \times N}$ denotes the inverse of the singular values $\Sigma$. After obtaining the $\tilde{H}$ matrix, the eigenvalue analysis is performed to obtain $\omega$ eigenvectors and $\sigma$ a diagonal matrix containing the corresponding eigenvalues.

$$\tilde{H}\omega = \sigma\omega \quad (7)$$

It is known that the eigenvalues of $\tilde{H}$ approximate some of the eigenvalues of the full system $A$. The associated eigenvectors of $H$ provide the coefficients for the linear combination that is necessary to express the dynamics within the image sequence basis.

The dynamic modes $\Psi$ are thus calculated as follows:

$$\Psi = P_2 V \Sigma^{-1} \omega \quad (8)$$

The complex eigenvalues $\sigma$ contain growth/decay rates and frequencies of the corresponding DMD modes [9], [10].

According to the authors in [8], DMD modes with Fourier frequencies $\mu_j$ is given by $\mu_j = \frac{\ln(\sigma_j)}{\delta t}$. Here, $\delta t$ is the time difference between the images and considered to be 1 in this study. The real part of $\mu_j$ regulates the growth or decay of the DMD modes, while the imaginary part of $\mu_j$ drives oscillations in the DMD modes. The frequencies near the origin (zero-modes) are interpreted as background (low-rank) portions of the given image sequence, and the Fourier frequencies bounded away from the origin are their sparse counterparts. Specifically, the parts in an image sequence that do not change in time, have an associated Fourier frequency at the origin of the complex plane with $\|\mu_j\| \approx 0$, which corresponds to the background model.

The methodology, as summarised above, allows us to obtain the DMD mode that reveals the background. Although this DMD mode has a dimension of $(3 \times mn) \times 1$ which can be translated back to the original image space of size m×n×3, it is nonetheless not a properly normalised image. As a result, the DMD mode needs to be calibrated in order to produce a proper image. To this end, we normalise the DMD mode to the range of [0 1] and then apply a color transfer from the statistically calculated mode image using the method proposed by Reinhard et al. [21]. In our preliminary experiments, we notice that the background models obtained from the statistically calculated mode and WS2006 [5] have equal CQM values and qualitatively look similar.

*C. Implementation*

Recall our methodological pipeline from Figure 1. For each of the color image sequences in the dataset consisting of N images, R, G and B channels are separated and converted to column vectors. These vectors are then vertically concatenated to form a data matrix as shown in Section II (Equation 1). DMD is then applied on the data matrix to obtain $N - 1$ dynamic modes for N images in the sequence. Fourier frequencies are calculated on the dynamic eigenvalues and the mode with the Fourier frequency $\approx 0$ is selected as the background mode. The three channels of that particular mode are first normalised and then combined to produce a color background initialised image for a given input image sequence. In order to retain the colormap from the original sequence, Reinhard *et al.'s* [21] method of color transfer is used. We refer the background produced from DMD as 'DMD' and background produced after color transfer as 'DMD_CT'.

*D. Evaluation metric*

Let GT be the groundtruth image and CB be the DMD computed background. We have adopted the image Color Quality Measure (CQM) from [1] as a metric for evaluating the performance of our technique: CQM is a recently proposed metric which is based on reversible luminance and chrominance (YUV) color transformation and peak signal-to-noise ratio (PSNR) measure [22]. The units of CQM are denoted in decibels db. The higher the CQM value, the better is the background model.

The evaluation is performed through the Matlab codes provided via the SBI dataset website[2].

### III. RESULTS & DISCUSSIONS

DMD background mode considers only those parts in an image sequence that does not change in time as a background model. When this condition holds, DMD performs better than Median, WS2006 and Photomontage methods. For example, in the case of Snellen sequence, the foreground leaves continuously move throughout the sequence and occupy most of the scene. Therefore, DMD is able to consider only those parts in the sequence that has not changed in time as its background model, while the aforementioned methods failed to remove the leaves from their background models as shown in Figure 2 (c, e & g). This is also confirmed by the quantitative results as DMD (36.17) and DMD_CT (36.85) outperformed the Median (36.07), WS2006 (24.99) and Photomontage (26.92) methods. Another example where the condition for DMD holds is the sequences of Highways, which generally reveal more than 50% of the background and most importantly the vehicles, do not remain stationary anywhere in the scene throughout the sequence. Therefore, DMD on these sequences could obtain a perfect background model by eliminating the foreground moving vehicles. Although, the aforementioned methods have been successful in obtaining a perfect background model, quantitatively DMD outperformed these methods as shown in Table I. Similarly, on the People&Foliage sequence DMD produced better results when compared to Median, SC-SOBS and WS2006. On the Foliage, DMD both quantitatively and qualitatively outperformed Median by eliminating the greenish halos produced due to the moving leaves in the foreground.

---
[2] http://sbmi2015.na.icar.cnr.it/SBIdataset.html

On the contrary, when the condition is violated, block based algorithms such as RS2011 may work better. For example, on the CaVignal sequence, DMD_CT and DMD qualitatively could not outperform the benchmark methods. This is due to the fact that, the man standing in the left of the sequence covers the scene for the first 60% of the sequence before he starts walking and again stands on the right for the last 10% of the sequence. Therefore, holding to the condition, DMD successfully eliminated the images when the person walks but not when he remains standing. The only effective solution to this problem is to track the object across the video sequence. However, since DMD does not rely on such a prior knowledge, its failure in this case is not unexpected. Nevertheless, quantitatively, DMD_CT (44.87) outperformed the rest of the methods except for RS2011 (52.59). Similarly for the Hall&Monitor sequence, a walking man in the corridor occupied the same scene region for more than 65% of the image sequence. Therefore, DMD included a ghost artefact of the man in its background model. Although methods such as WS2006, Photomontage and RS2011 were able to obtain a perfect background model, interestingly they have not performed well quantitatively comparing with DMD.

Prior to comparing the overall average CQM value across the seven sequences with the methods presented in [1], we first compare the results produced from DMD (Figure 2 (h)) and DMD_CT (Figure 2 (i)). After transferring the colormap from the original image sequence to the 'DMD' background initialised image, we see that the intensity values in 'DMD_CT' are adjusted to the original sequence's color format. For instance, on the HighwayI and HighwayII sequence, the background model produced from DMD had a different color intensity compared to the groundtruth, but when the colormap is transferred, the background model produced by DMD_CT acquired the original sequence's color format. Quantitative results show the background models produced from DMD_CT have better CQM values compared to the background models produced from DMD (except for the Hall&Monitor sequence, but the difference is marginal), as shown in Table I. On average, across all seven sequences DMD_CT (42.82) performed better than the Median (39.00), WS2006 (39.87) and Photomontage (42.82) methods and stands at rank 3 as shown in Table I.

## IV. CONCLUSIONS

In this paper, we have introduced a technique using DMD to obtain the background model in the color domain. The significance of our technique is demonstrated on a publicly available SBI dataset. When DMD condition holds, i.e., "DMD background mode considers only those parts in an image sequence that does not change in time as a background model", it performs better than Median, WS2006 and Photomontage. Contrarily, when the condition is violated, block-based algorithm such as RS2011 may work better. Since DMD_CT (with color calibration) produces visibly higher quality image than DMD, we recommend that DMD_CT be used when the condition is true. Our results in this paper show that DMD can successfully obtain a colored background model.

ACKNOWLEDGMENT

The funding for this work has been provided by the Department of Computer Science and the Centre for Vision, Speech and Signal Processing (CVSSP) - University of Surrey. "ST" would like to thank Dr. Simon Bull for his valuable comments.

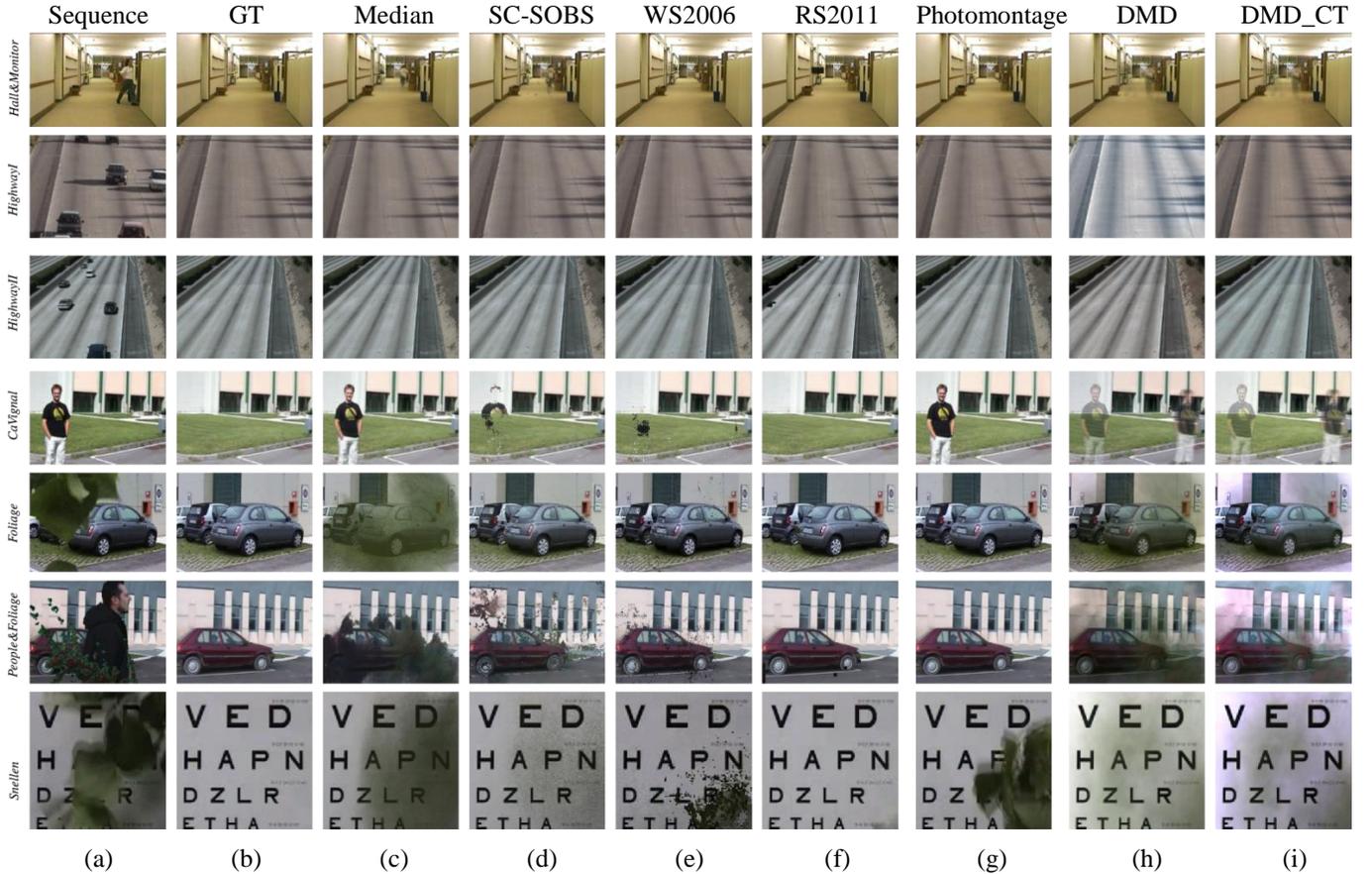

Fig. 2. (a) Exemplar frames corresponding to {295, 0, 0, 0, 261, 10, and 0} from seven different sequences of the SBI dataset. (b) Corresponding GT's for the sequences. (c-g) Background models generated by benchmarked methods: (c) Median, (d) SC-SOBS, (e) WS2006, (f) RS2011, (g) Photomontage. (h) DMD and (i) DMD_CT. Both DMD and DMD_CT are the proposed methods in this paper. The images {a-g} are taken from [1].

TABLE I. QUANTITATIVE COMPARISON OF CQM VALUES FOR DMD & DMD_CT WITH THE BENCHMARK METHODS [1].

|  | CaVignal | Foliage | Hall&Monitor | HighwayI | HighwayII | People&Foliage | Snellen | Avg(CQM) | Rank |
|---|---|---|---|---|---|---|---|---|---|
| SC-SOBS [4] | 42.27 | 39.11 | 43.19 | 65.58 | 54.38 | 35.37 | 44.75 | 46.38 | 1 |
| RSL2011 [6] | 52.59 | 43.10 | 35.08 | 38.00 | 51.98 | 37.06 | 50.26 | 44.01 | 2 |
| DMD_CT | 44.87 | 34.65 | 42.39 | 55.77 | 46.95 | 31.97 | 36.85 | 41.92 | 3 |
| Photomontage [7] | 32.06 | 45.61 | 41.73 | 59.03 | 35.08 | 47.15 | 26.92 | 41.08 | 4 |
| DMD | 40.17 | 34.39 | 42.53 | 52.36 | 45.07 | 28.43 | 36.17 | 39.87 | 5 |
| Median | 33.14 | 28.73 | 62.57 | 42.67 | 42.32 | 27.50 | 36.07 | 39.00 | 6 |
| WS2006 [5] | 37.06 | 34.98 | 40.09 | 56.91 | 40.51 | 31.38 | 24.99 | 37.99 | 7 |